\documentclass[11pt]{article}

\usepackage{amsmath,epsfig,amssymb,amsthm}
\usepackage[boxed]{algorithm2e}
\usepackage{graphicx}
\usepackage{amstext}
\usepackage{amsthm}
\usepackage{flushend}

\newcommand{\bm}[1]{\boldsymbol{\mathrm{#1}}}

\def\Tr{\mathrm{T}}

\newcommand\tr{\mathrm{tr}\,}

\renewcommand{\baselinestretch}{1.2}

\newcommand{\isfulltr}{}

\begin{document}

\title{Spectral descriptors for deformable shapes}

\author{A. M. Bronstein
\thanks{School of Electrical Engineering, Tel Aviv University, Tel Aviv 69978, Israel.
Email: bron@eng.tau.ac.il. This work was supported by the Israeli Science Foundation Grant and the German-Israeli Foundation.}
}
\maketitle

\begin{abstract}
Informative and discriminative feature descriptors play a fundamental role in deformable shape analysis. For example, they have been successfully employed in correspondence, registration, and retrieval tasks.
In the recent years, significant attention has been devoted to descriptors obtained from the spectral decomposition of the Laplace-Beltrami operator associated with the shape. Notable examples in this family are the heat kernel signature (HKS) and the wave kernel signature (WKS). Laplacian-based descriptors achieve state-of-the-art performance in numerous shape analysis tasks; they are computationally efficient, isometry-invariant by construction, and can gracefully cope with a variety of transformations.
In this paper, we formulate a generic family of parametric spectral descriptors. We argue that in order to be optimal for a specific task, the descriptor should take into account the statistics of the corpus of shapes to which it is applied (the ``signal'') and those of the class of transformations to which it is made insensitive (the ``noise''). While such statistics are hard to model axiomatically, they can be learned from examples. Following the spirit of the Wiener filter in signal processing, we show a learning scheme for the construction of optimal spectral descriptors and relate it to Mahalanobis metric learning.
The superiority of the proposed approach is demonstrated on the SHREC'10 benchmark. 
\end{abstract}

\section{Introduction}

The notion of a \emph{feature descriptor} is fundamental in shape analysis.
A feature descriptor assigns each point on the shape a vector in some single- or multi-dimensional feature space representing the point's local and global geometric properties relevant for a specific task. This information is subsequently used in higher-level tasks: for example, in shape matching descriptors are used to establish an initial set of potentially corresponding points \cite{gelfand2005rgr,wang2011matching}; in shape retrieval a global shape descriptor is constructed as a bag of ``geometric words'' expressed in terms of local feature descriptors \cite{mitra2006pfs,bronstein2011shape}; segmentation algorithms rely on the similarity or dissimilarity between feature descriptors to partition the shape into stable and meaningful parts \cite{skraba2010persistence}.

When constructing or choosing a feature descriptor, it is imperative to answer two fundamental questions: which shape properties the descriptor has to capture, and to which transformations of the shape it shall remain invariant.

\subsection{Previous work}

 Early research on feature descriptors focused mainly on invariance under global Euclidean transformations (rigid motion). Classical works in this category include the shape context \cite{belongie2000shape} and spin image \cite{johnson1999usi} descriptors, as well as integral volume descriptors \cite{manay2004iis,gelfand2005rgr} and multiscale local features \cite{pauly2003msf} just to mention a few out of many.

In the past decade, significant effort has been invested in extending the invariance properties to non-rigid deformations. Some of the classical rigid descriptors were extended to the non-rigid case by replacing the Euclidean metric with its geodesic counterpart \cite{hamza2003geodesic,elad2003bending}. Also, the use of conformal factors has been proposed \cite{lipman2009mobius}. Being intrinsic properties of a surface, both are independent of the way the surface is embedded into the ambient Euclidean space and depend only on its metric structure. This makes such descriptors invariant to inelastic bending transformations.
However, geodesic distances suffer from strong sensitivity to topological noise, while conformal factors, being a local quantity, are influenced by geometric noise. Both types of noise, virtually inevitable in real applications, limit the usefulness of such descriptors.

Recently, a family of intrinsic geometric properties broadly known as \emph{diffusion geometry} has become growingly popular.
The studies of diffusion geometry are based on the theoretical works by Berard \emph{et al.} \cite{berard1994embedding} and later by Coifman and Lafon \cite{coifman2006diffusion} who suggested to use the eigenvalues and eigenvectors of the Laplace-Beltrami operator associated with the shape to construct invariant metrics known as diffusion distances. These distances as well as other diffusion geometric constructs have been show significantly more robust compared to their geodesic counterparts \cite{memoli2009spectral,bronstein2010gromov}.
Diffusion geometry offers an intuitive interpretation of many shape properties in terms of spacial frequencies and allows to use standard harmonic analysis tools. Also, recent advances in the discretization of the Laplace-Beltrami operator bring forth efficient and robust numerical and computational tools.

These methods were first explored in the context of shape processing by L\'{e}vy \cite{lévy2006laplace}. Several attempts have also been made to construct feature descriptors based on diffusion geometric properties of the shape. Rustamov \cite{rustamov2007laplace} proposed to construct the \emph{global point signature} (GPS) feature descriptors by associating each point with an $\ell^2$ sequence based on the eigenfunctions and the eigenvalues of the Laplacian, closely resembling a diffusion map \cite{coifman2006diffusion}. A major drawback of such a descriptor was its ambiguity to sign flips of each individual eigenfunction (or, in the most general case, to rotations and reflections in the eigenspaces corresponding to each eigenvalue).

A remedy was proposed by Sun \emph{et al.} who in their influential paper \cite{sun2009concise} introduced the \emph{heat kernel signature} (HKS), based on the fundamental solutions of the heat equation (heat kernels). In \cite{aubry2011wave}, another physically-inspired descriptor, the \emph{wave kernel signature} (WKS) was proposed as a solution to the excessive sensitivity of the HKS to low-frequency information. As of today, these descriptors achieve state-of-the-art performance in many deformable shape analysis tasks \cite{bronstein2010shrec_corr,bronstein2010shrec_ret}.

\subsection{Contribution}

In this paper, we remain within the diffusion geometric framework and propose a generic family of spectral feature descriptors that generalize both the HKS and the WKS. We analyze both descriptors within this framework pointing to their advantages and drawbacks, and enumerate a list of desired properties a descriptor should have.

We argue that in order to construct a good task-specific spectral descriptor, one has to be in the position of defining the spectral content of the geometric ``signal'' (i.e., the properties distinguishing different classes of shapes from each other) and the ``noise'' (i.e., the changes of the latter properties due to the deformations the shapes undergo). Both are functions of the corpus of data of interest, and the transformations invariance to which is desired. While it is notoriously difficult to characterize such properties analytically, we propose to learn them from examples in a way resembling the construction of a Wiener filter that passes frequencies containing more signal than noise, while attenuating those where the noise covers the signal.

This study was in part inspired by the insightful paper by Auby \emph{et al.} \cite{aubry2011wave}, and in part is a continuation of \cite{aflalo2011wiener} where we attempted to construct optimal diffusion metrics. However, since diffusion metrics are characterized by a single frequency response, the attempt had a modest success. On the other hand, vector-valued feature descriptors allowing for multiple frequency response functions have, in our opinion, more potential. This paper does not intend to exhaust this potential, but merely to explore a part of it.

The rest of the paper is organized as follows: In Section~2 we introduce the mathematical notation of the Laplace-Beltrami operator and its spectrum and briefly overview the state-of-the-art descriptors based on its properties. In Section 3, we indicate several drawbacks of these descriptors and analyze the properties a good descriptor should satisfy. We present a spectral descriptor generalizing the heat and the wave kernel signatures, and show an approach for learning its optimal task-specific parameters from examples. Relation to metric learning is highlighted.
In Section~4, the superiority of the proposed learnable descriptor over the fixed ones is shown experimentally on the SHREC'10 non-rigid correspondence benchmark.
Finally, Section~5 concludes the paper.

Since the figures visualizing the experiments in Section~4 are relatively self-explanatory, we decided to incorporate them in the flow as illustrations to the phenomena discussed in the paper even before the exact experimental setting are detailed.

\section{Spectral descriptors}


We model a shape as a compact two-dimensional manifold $X$, possibly with a boundary $\partial X$.
The manifold is endowed with a Riemannian metric defined as a local inner product $\langle \cdot, \cdot \rangle_x$
 on the tangent plane $T_x X$ at each point $x\in X$.
Given a smooth scalar field $f$ on the manifold, its gradient $\mathrm{grad}\,f$ is the vector field
satisfying $f(x+dr) = f(x) + \langle \mathrm{grad}\, f(x), dr \rangle_x$ for every infinitesimal tangent vector $dr \in T_x X$.
The inner product $\langle \mathrm{grad}\, f(x), v \rangle_x$ can be interpreted as the directional derivative of $f$ in the direction $v$.
A directional derivative of $f$ whose direction at every point is defined by a vector field $V$ on the manifold is called the Lie derivative of $f$ along $V$.
The Lie derivative of the manifold volume (area) form along a vector field $V$ is called the divergence of $V$, $\mathrm{div}\,V$.
The negative divergence of the gradient of a scalar field $f$,
$\Delta f = -\mathrm{div}\, \mathrm{grad}\,f$, is called the \emph{Laplacian} of $f$. The operator $\Delta$ is called the \emph{Laplace-Beltrami} operator, and it
generalizes the standard notion of the Laplace operator to manifolds.
Note that we define the Laplacian with the negative sign to conform to the computer graphics and computational geometry convention.
%

\subsection{Laplacian spectrum and Shape DNA}

Being a positive self-adjoint operator, the Laplacian admits an eigendecomposition
\begin{eqnarray}
\Delta \phi = \nu \phi
\label{eq:helmoltz}
\end{eqnarray}
with non-negative eigenvalues $\nu$ and corresponding orthogonormal eigenfunctions $\phi$. Furthermore, due to the assumption that our domain is compact,
the spectrum is discrete, $0 = \nu_1 < \nu_2 < \cdots $.

In physics, (\ref{eq:helmoltz}) is known as the \emph{Helmohltz equation} representing the spatial component of the wave equation. Thinking of our domain as of a vibrating membrane (with appropriate boundary conditions), the $\phi_k$'s can be interpreted as natural vibration modes of the membrane, while the $\nu_k$'s assume the meaning of the corresponding vibration frequencies. In fact, in this setting the eigenvalues have inverse area or squared spatial frequency units.

This physical interpretation leads to a natural question whether the eigenvalues of the Laplace-Beltrami operator fully determine the shape of the domain. The essence of this question was beautifully captured by Mark Kac as \emph{``can one hear the shape of the drum?''} \cite{kac1966can}. Unfortunately, the answer to this question is negative as there exist isospectral manifolds that are not isometric.
The exact relation between the latter two classes of shapes is unknown, but it is generally believed that most isospectral manifolds are also isometric.
Based on this belief, Reuter \emph{et al.} \cite{reuter2006laplace} proposed to use truncated sequences of the Laplacian eigenvalues as isometry-invariant shape descriptors, dubbed by the authors as \emph{shape DNA}.

\subsection{Heat kernel signature}

The Laplace-Beltrami operator plays a central role in the \emph{heat equation} describing diffusion processes on manifolds.
In our notation, the heat equation can be written as
\begin{eqnarray}
\left (\Delta + \frac{\partial }{\partial t}\right) u(x,t)  =  0
\end{eqnarray}
where $u(x,t)$ is the distribution of heat on the manifold at point $x$ at time $t$. The initial condition is
some initial heat distribution $u_0(x)$ at time $t=0$, and boundary conditions are applied in case the manifold has a boundary.

The solution of the heat equation at time $t$ can be expressed as the application of the \emph{heat operator}
\begin{eqnarray}
u(x,t) = \int h_t(x,y) u_0(y) da(y)
\label{eq:heat}
\end{eqnarray}
to the initial distribution. The kernel $h_t(x,y)$ of this integral operator is called the \emph{heat kernel} and it corresponds to the solution
of the heat equation at point $x$ at time $t$ with the initial distribution being a delta function at point $y$. From the signal processing perspective, the heat kernel
can be interpreted as a non shift-invariant ``impulse response''. It also describes the amount of heat transferred from point $x$ to point $y$ after time $t$, as well as the transition probability density from point $x$ to point $y$ by a random walk of length $t$.

According to the spectral decomposition theorem, the heat kernel can be expressed as
\begin{eqnarray}
h_t(x,y) = \sum_{k \ge 1} \exp(-\nu_k t) \phi_k(x) \phi_k(y),
\end{eqnarray}
where $\exp(-\nu t)$ can be interpreted as its ``frequency response'' (note that with a proper selection of units in (\ref{eq:heat}), the eigenvalues $\nu_k$ assume inverse time or frequency units). The bigger is the time parameter, the lower is the cut-off frequency of the low-pass filter described by this response and, consequently, the bigger is the support of $h_t$ on the manifold.
The quantity
\begin{eqnarray}
h_t(x,x) = \sum_{k \ge 1} \exp(-\nu_k t) \phi^2_k(x),
\end{eqnarray}
sometimes referred to as the \emph{autodiffusivity function} \cite{sharma2010shape}, describes the amount of heat remaining at point $x$ after time $t$.
Furthermore, for small values of $t$ is it related to the manifold curvature
according to
\begin{eqnarray}
h_t(x,x) = \frac{1}{4 \pi t} + \frac{K(x)}{12 \pi} + \mathcal{O}(t),
\end{eqnarray}
where $K(x)$ denotes the Gaussian (in general, sectional) curvature at point $x$.

In \cite{sun2009concise}, Sun \emph{et al.} showed that under mild technical conditions, the sequence $\{ h_t(x,x) \}_{t>0}$ contains \emph{full} information about the metric of the manifold. The authors proposed to associate each point $x$ on the manifold with a vector
\begin{eqnarray}
\bm{p}(x) = \left( h_{t_1}(x,x), \dots, h_{t_n}(x,x) \right)^\Tr,
\end{eqnarray}
of the autodiffusivity functions sampled at some finite set of times $t_1,\dots,t_n$. The authors dubbed such a feature descriptor as the \emph{heat kernel signature}.
In \cite{bronstein2011shape}, an HKS-based bag-of-features approach was introduced under the name of Shape Google and was shown to achieve state-of-the-art results in deformable shape retrieval.
In \cite{Iasonas}, a scale-invariant version of the HKS was proposed, and \cite{raviv2010volumetric} extended the descriptor to volumes.

\begin{figure}[tbp]
    \begin{center}
\ifx\isfulltr\undefined
   \includegraphics[width=1\columnwidth]{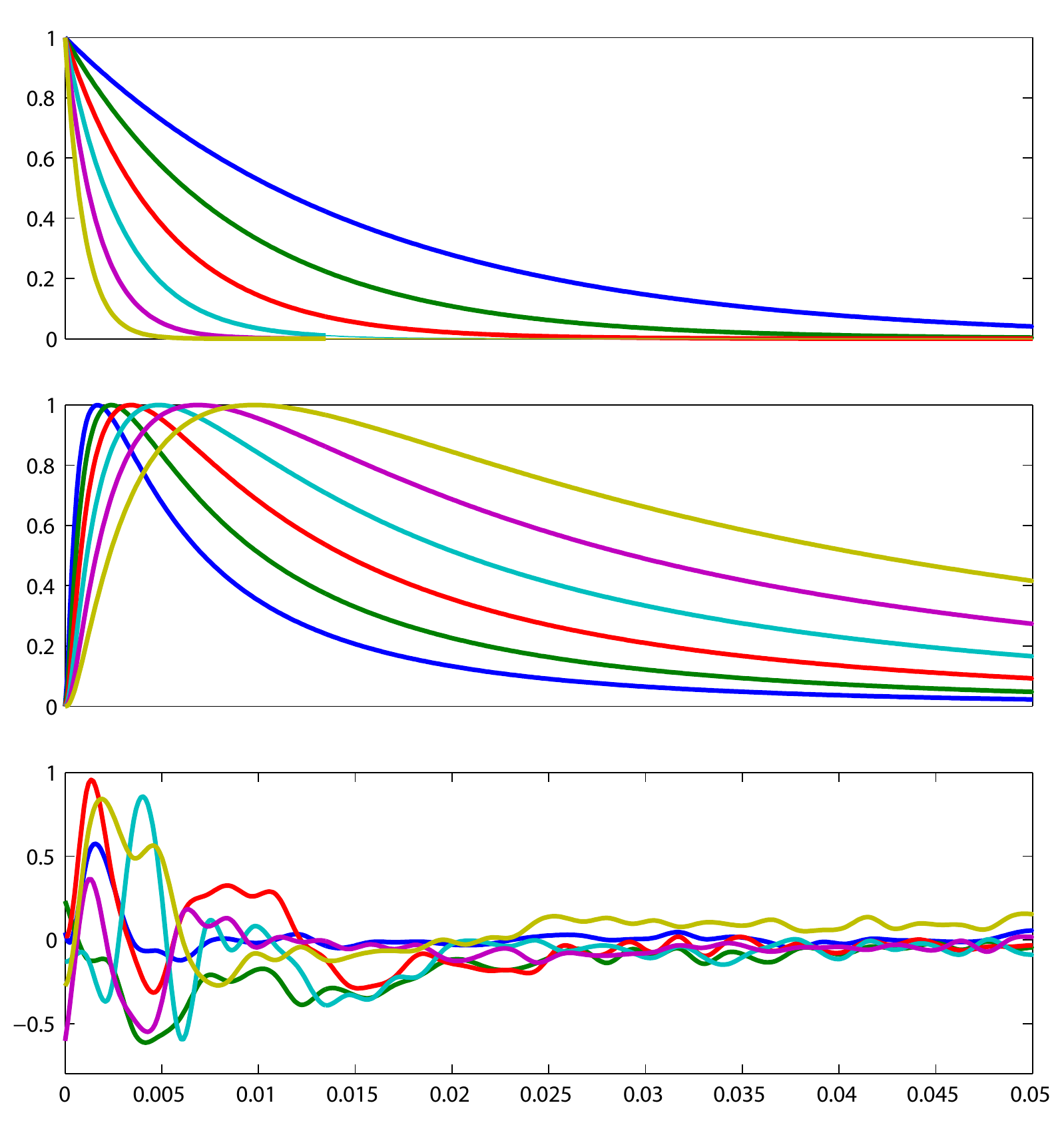}
\else
   \includegraphics[width=.8\columnwidth]{figures/kernels-eps-converted-to.pdf}
\fi
   \caption{\label{fig:kernels} Examples of (unnormalized) kernels used for the computation of the heat kernel (first row), wave kernel (second row), and trained optimal kernel (last row) descriptors.  }
    \end{center}
\end{figure}

\begin{figure*}
    \begin{center}
\ifx\isfulltr\undefined
   \includegraphics[width=0.8\linewidth]{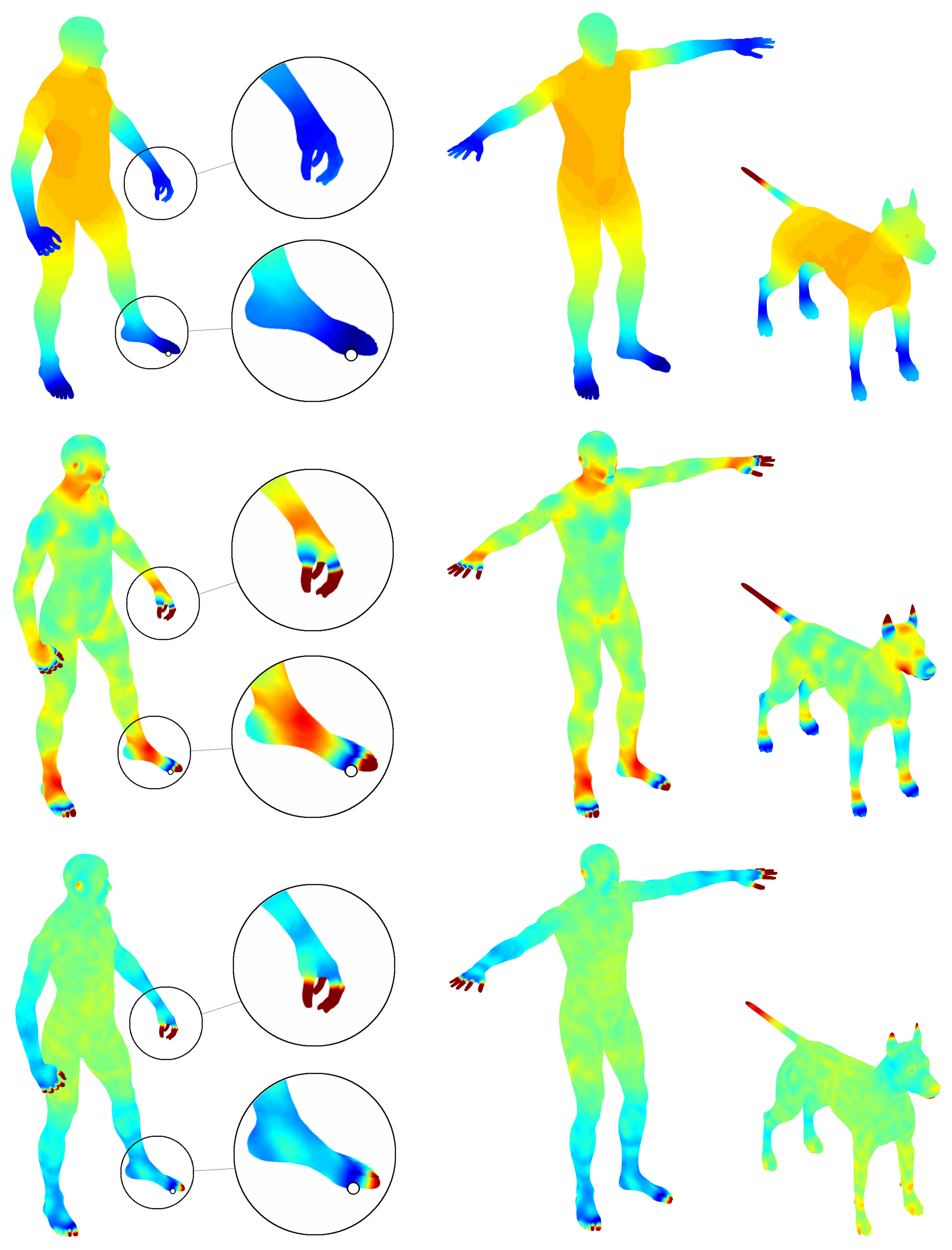}
\else
   \includegraphics[width=1\linewidth]{figures/dist.png}
\fi    
   \caption{\label{fig:distance} Normalized Euclidean distance between the descriptor at a reference point on the left foot (white dot in the leftmost column) and descriptors computed at rest of the points of the same shape (left column), its approximate isometry (middle column), and a distinct shape (right column). Twelve-dimensional descriptors based on the heat kernel (first row), wave kernel (second row), and trained optimal kernel (last row) are shown. Dark blue stands for small distance; red represents large distance.}
    \end{center}
\end{figure*}

Despite its success, the heat kernel descriptor suffers from several drawbacks. First, being a collection of low-pass filters (Figure~\ref{fig:kernels}, top), the descriptor is dominated by low frequencies conveying information mostly about the global structure of the shape. While being important to discriminate between distinct shapes (which usually differ greatly at coarse scales), this emphasize of low frequencies damages the ability of the descriptor to precisely localize features. This phenomenon can be observed in Figure~\ref{fig:distance} (top). In fact, the distance between HKS computed at a point $x$ and HKS of neighboring points increases slowly, while for good localization a steeper increase is required.

\subsection{Wave kernel signature}

A remedy to the poor feature localization of the heat kernel descriptor was proposed by Aubry \emph{et al.} \cite{aubry2011wave}.
The authors proposed to replace the heat diffusion model that gives rise to the HKS by a different physical model in which one evaluates the probability of a quantum particle with a certain energy distribution to be located at a point $x$.
The behavior of a quantum particle on a surface is governed by the Schr\"{o}dinger equation
\begin{eqnarray}
\left (i\Delta + \frac{\partial }{\partial t}\right) \psi(x,t) = 0
\end{eqnarray}
where $\psi(x,t)$ is the complex wave function. 
Despite an apparent similarity to the heat equation, the multiplication of the Laplacian by the complex unity in the Schr\"{o}dinger equation has a dramatic impact on the dynamics of the solution.
Instead of representing diffusion, $\psi$ now has oscillatory behavior.

Let us assume that the quantum particle has an initial energy distribution $f(e)$. Since energy is directly related to frequency, we will use $f(\nu)$ instead in order to stick to the previous notation.
The solution of the Schr\"{o}dinger equation can then be expressed in the spectral domain as \cite{aubry2011wave}
\begin{eqnarray}
\psi(x,t) = \sum_{k\ge 1} \exp{(i\nu_k t)} f(\nu_k) \phi_k(x)
\end{eqnarray}
(note the complex unity in the exponential!). The probability to measure the particle at a point $x$ at time $t$ is given by $|\psi(x,t)|^2$.
By integrating over all times, the average probability
\begin{eqnarray}
p(x) = \lim_{T \rightarrow \infty} \frac{1}{T} \int_0^T |\psi(x,t)|^2 dt
= \sum_{k\ge 1} f^2(\nu_k) \phi_k^2(x).
\label{eq:prob}
\end{eqnarray}
to measure the particle at a point $x$ is obtained. Note that the probability depends on the initial energy distribution $f$.

Aubry \emph{et al.} considered a family of log-normal energy distributions
\begin{eqnarray}
f_e(\nu) \propto \exp\left( -\frac{(\log e - \log \nu)^2}{2 \sigma^2} \right)
\label{eq:fe}
\end{eqnarray}
centered around some mean log energy $\log e$ with variance $\sigma^2$ (again, we allow ourselves a certain abuse of the physics and treat energy and frequency as synonyms).
This particular choice of distributions is motivated by a perturbation analysis of the Laplacian spectrum \cite{aubry2011wave}.

Fixing the family of energy distributions, each point on the surface is associated with a \emph{wave kernel signature} of the form
\begin{eqnarray}
\bm{p}(x) = \left( p_{e_1}(x), \dots, p_{e_n}(x) \right)^\Tr,
\end{eqnarray}
where $p_e(x)$ is the probability to measure a quantum particle with the initial energy distribution $f_e(\nu)$ at point $x$. 
The authors use logarithmically sampled $e_1,\dots,e_n$.

The WKS descriptor resembles the HKS in the sense that it can also be thought of as an application of a set of filters with the frequency responses $f^2_e(\nu)$. However, unlike
the HKS that uses low-pass filters, the responses of the WKS are \emph{band-pass} (Figure~\ref{fig:kernels}, middle). This reduces the influence of the low frequencies and allows better separation of frequency bands across the descriptor dimensions.
As the result, the wave kernel descriptor exhibits superior feature localization (Figure~\ref{fig:distance}, middle).

\section{Spectral descriptor learning}

Despite their beautiful physical interpretation, both the heat and wave kernel descriptors suffer from several drawbacks.

The fact that the WKS deemphasizes large-scale features contributes to its higher \emph{sensitivity} (i.e., the ability to identify positives). This property is crucial in matching problems, where a small set of candidate matches on one shape is found for a collection of reference points on the other. The ability to produce a correct match within a small set of best matches (high true positive rate at low false positive rate) greatly increases the performance of correspondence algorithms.

On the other hand, by emphasizing global features HKS has higher \emph{specificity} (i.e., the ability to identify negatives). This property is related to discriminativity, that is, the ability of  the descriptor to distinguish between a shape and other classes of distinct shapes. High discriminativity is important in retrieval applications, and the performance of the descriptor at low false negative rates has a big impact on retrieval algorithms based on it.
Both phenomena are visualized in Figure~\ref{fig:roc}.
While it is impossible to maximize both the sensitivity and the specificity, a good descriptor is expected to have both reasonably high.

Another drawback of both the heat and wave kernel descriptors is the fact that the frequency responses forming their elements have significant overlaps. As the results, the descriptor has redundant dimensions.
Finally, both the heat and wave kernel signatures are only invariant to truly isometric deformations of the shape (and can be also made scale-invariant using the scheme proposed in \cite{Iasonas}). Deformations that real shapes undergo frequently deviate from this model, and it is unclear how they influence the performance of the HKS and WKS.

We believe that many real-world deformations affect different frequencies differently. At the same time, the geometric features that allow to localize a point on a shape or to distinguish a shape from other shapes also depend differently on different frequencies. Emphasizing information-carrying frequencies while attenuating noise-carrying ones is a classical idea in signal and is the underlying principle of Wiener filtering \cite{wiener1964extrapolation}.

\begin{figure*}[tbp]
    \begin{center}
\ifx\isfulltr\undefined
   \includegraphics[width=0.49\linewidth]{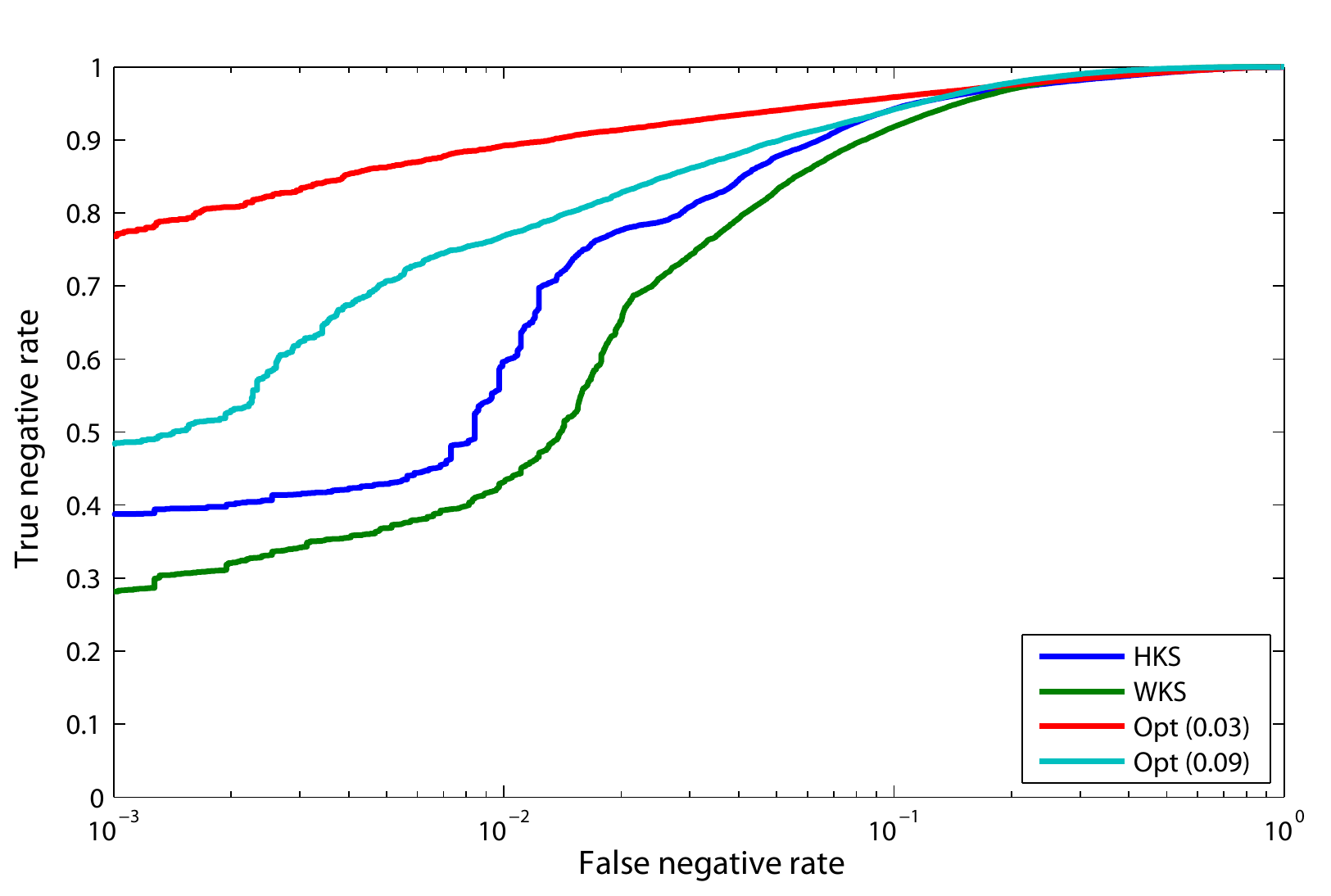} \includegraphics[width=0.49\linewidth]{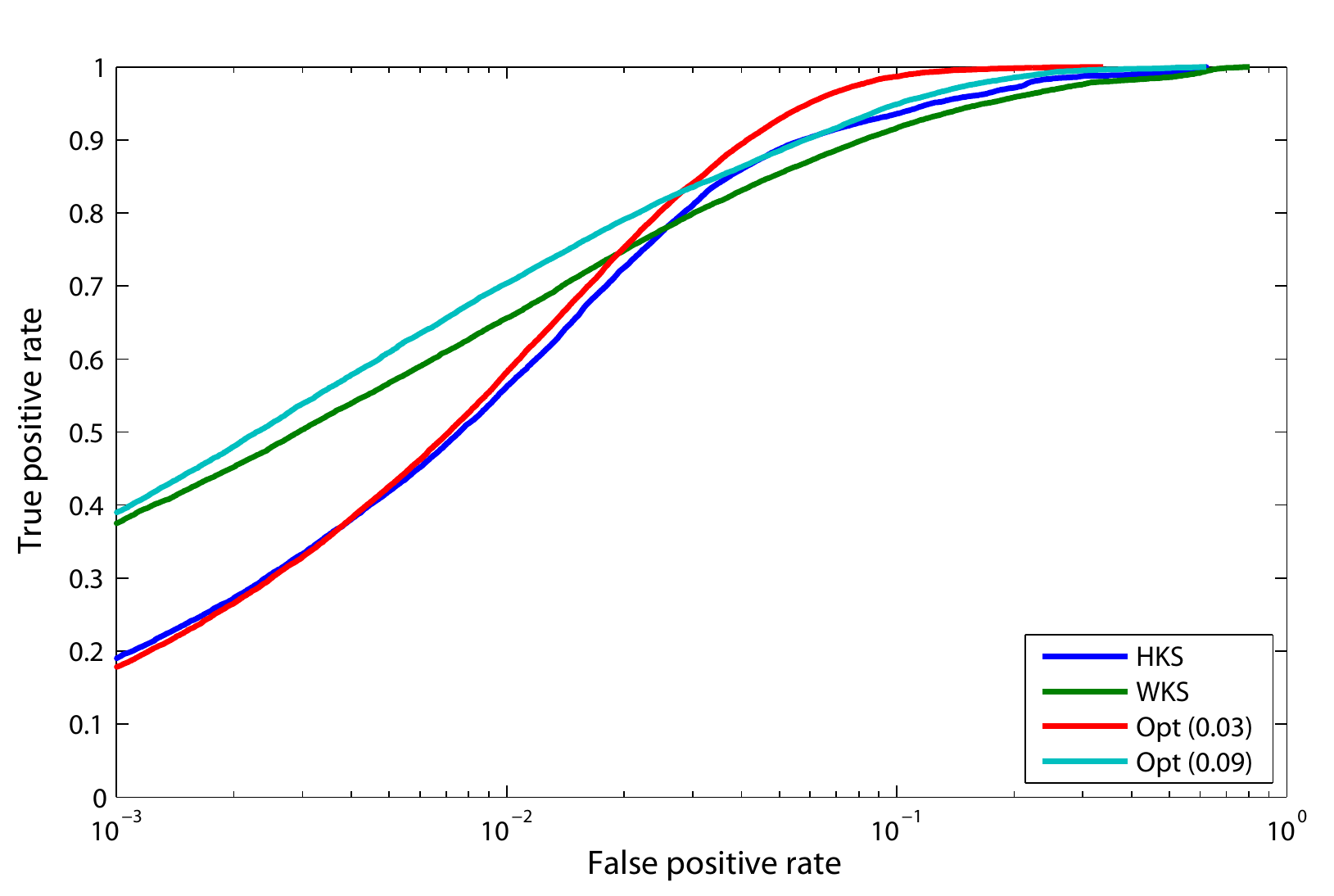}
   \caption{\label{fig:roc} ROC curves of different spectral descriptors when matching points of a shape to itself. A positive match is considered within a geodesic ball of $1\%$ of the shape diameter. Bilaterally symmetric matches are also considered positives. Two regions of the ROC curve are emphasized: the performance of the descriptors for low false negative rate (left), and low false positive rate (right). The former case is important to be able to discriminate between different shapes in shape retrieval applications, while the latter is required for establishing accurate correspondence between two shapes.   }
\else
   \includegraphics[width=0.8\linewidth]{figures/roc-fn-eps-converted-to.pdf}\\
   \includegraphics[width=0.8\linewidth]{figures/roc-fp-eps-converted-to.pdf}
   \caption{\label{fig:roc} ROC curves of different spectral descriptors when matching points of a shape to itself. A positive match is considered within a geodesic ball of $1\%$ of the shape diameter. Bilaterally symmetric matches are also considered positives. Two regions of the ROC curve are emphasized: the performance of the descriptors for low false negative rate (top), and low false positive rate (bottom). The former case is important to be able to discriminate between different shapes in shape retrieval applications, while the latter is required for establishing an accurate correspondence.   }
\fi
    \end{center}
\end{figure*}

\subsection{Desired properties}

This observation leads us to the main contribution of this paper: we propose to construct a collection of frequency responses forming an optimal spectral descriptor.
In order to be useful, such a  descriptor should satisfy the following properties:
\begin{enumerate}

\item \emph{Localization}: a small displacement of a point on the manifold should greatly affect the descriptor computed at it.

\item \emph{Sensitivity}: when a point on a shape is queried against another similar shape, a small set of best matches of the descriptor should contain a correct match with high probability.

\item \emph{Discriminativity}: the descriptor should be able to distinguish between shapes belonging to different classes. 

\item \emph{Invariance}: the descriptor should be invariant or at least insensitive to a certain class of transformations that the shape may undergo.

\item \emph{Efficiency}: the descriptor should capture as much information as possible within as little number of dimensions as possible.

\end{enumerate}
The localization and sensitivity properties are important for matching tasks, while in order to be useful in shape retrieval tasks, the descriptor should have the discriminativity property.
However, discriminativity is data-dependent: a descriptor can be discriminative on one corpus of data, while non-discriminative on another. While it is generally impractical to model classes of shapes axiomatically, machine learning offers an easy alternative of inferring them from training data.

By construction, spectral descriptors are isometry invariant. However, other invariance properties are usually hard to achieve and even harder to model for realistic transformations. We will therefore stick to learning in order to achieve invariance on examples of transformations the training shapes undergo.

\subsection{Parametrization}

We are interested in descriptors of the form
\begin{eqnarray}
\bm{p}(x) = \sum_{k \ge 1} \bm{f}(\nu_k) \phi_k^2(x),
\label{eq:descriptor}
\end{eqnarray}
parameterized by a vector $\bm{f}(\nu) = (f_1(\nu),\dots,f_n(\nu))^\Tr$ of frequency responses. Both the HKS and the WKS are particular cases of this general form.
Unlike both heat and wave kernels that are strictly positive, we will allow $\bm{f}(\nu)$ assume negative values.

Since the responses $\bm{f}(\nu)$ are the design variables of the descriptor, they have to be parametrized with a finite set of parameters. The same parameters have to be compatible with any shape, even though different shapes differ in the set of eigenvalues $\{ \nu_k \}$. In order to make the representation independent of a specific shape's eigenvalues, we fix a basis $\{b_1(\nu),\dots, b_m(\nu)\}$, $m>n$, spanning a sufficiently wide interval $[0,\nu_\mathrm{max}]$ of frequencies. This allows to express $\bm{f}(\nu)$ as
\begin{eqnarray}
\bm{f}(\nu) =
\bm{A}\bm{b}(\nu),
\label{eq:response-rep}
\end{eqnarray}
where $\bm{A}$ is the $n\times m$ matrix of coefficients representing the response using the basis functions $\bm{b}(\nu) = (b_1(\nu),\dots,b_m(\nu))^\Tr$.

Since the eigenvalues $\nu_k$ form a growing progression, we can truncate the series (\ref{eq:descriptor}) at $\nu_s \ge \nu_\mathrm{max}$. Substituting the representation (\ref{eq:response-rep}), we obtain
\begin{eqnarray}
\bm{p}(x) = \bm{A} (\bm{b}(\nu_1),\dots,\bm{b}(\nu_s))  \left( \begin{array}{c}
                       \phi_1^2(x) \\
                       \vdots \\
                       \phi_s^2(x)
                     \end{array}
\right) = \bm{A}\bm{g}(x)
\end{eqnarray}
where the $m\times 1$ vector $\bm{g}(x)$
with the elements
\begin{eqnarray}
g_j(x) = \sum_{k \ge 1} b_j(\nu_k) \phi_k^2(x)
\end{eqnarray}
captures all the shape-specific geometric information about the point $x$. For this reason, we refer to $\bm{g}$ as to the \emph{geometry vector} of a point.
Note that this representation is no more depends on a specific shape; the matrix of parameters $\bm{A}$ describes the same vector of frequency responses on any shape.

\subsection{Learning}

Let $\bm{g} = \bm{g}(x)$ be the geometry vector representing some point $x$; let $\bm{g}_+ = \bm{g}(x_+)$ be another geometry vector representing a point that is knowingly similar to $x$ (positive); and, finally, let  $\bm{g}_- = \bm{g}(x_-)$ represent a knowingly dissimilar point (negative).
We would like to select the matrix of parameters that maximizes the similarity of the descriptors $\bm{p} = \bm{A}\bm{g}$ and $\bm{p}_+ = \bm{A}\bm{g}_+$, and at the same time
minimizes the similarity between $\bm{p}$ and $\bm{p}_- = \bm{A}\bm{g}_-$. Using the $L^2$ norm as the similarity criterion, we obtain
\begin{eqnarray}
d^2_\pm &=& \| \bm{p} - \bm{p}_\pm \|^2 = \| \bm{A}(\bm{g} - \bm{g}_\pm) \|^2 \nonumber\\
&=& (\bm{g} - \bm{g}_\pm)^\Tr \bm{A}^\Tr \bm{A} (\bm{g} - \bm{g}_\pm).
\end{eqnarray}
In other words, the Euclidean distance between the descriptors translates into a Mahalanobis distance between the corresponding geometry vectors. The problem of finding the best positive-definite matrix $\bm{A}^\Tr \bm{A}$ defining the Mahalanobis metric is known as \emph{metric learning} and has been relatively well explored in the literature \cite{yang2006distance,weinberger2006distance,davis2007information}.

Here, we describe a simple yet efficient learning scheme explicitly addressing the desired properties we required from a good spectral descriptor.
We aim at finding a matrix $\bm{A}$ minimizing the Mahalanobis distance over the set of positive pairs, while maximizing it over the negative ones.
Note that the distance depends only on the differences between positive and negative pairs of vectors.
Taking expectation over all positive and negative pairs, we obtain \cite{strecha2011ldahash}
\begin{eqnarray}
\mathbb{E}(d^2_\pm) &=&  \mathbb{E}( \| \bm{p} - \bm{p}_\pm \|^2 ) = \mathbb{E} (\bm{e}^\Tr_\pm \bm{A}^\Tr \bm{A} \bm{e}_\pm) \nonumber\\
&=& \tr( \bm{A} \mathbb{E} (\bm{e}_\pm\bm{e}^\Tr_\pm)  \bm{A}^\Tr )
= \tr( \bm{A} \bm{C}_\pm  \bm{A}^\Tr ),
\end{eqnarray}
where $\bm{e}_\pm = \bm{g} - \bm{g}_\pm$, and $\bm{C}_\pm$ stands for the covariance matrix of the differences of positive and negative pairs of geometry vectors.
In practice, the expectations are replaced by averages over a representative set of difference vectors.

Our goal is to minimize $\mathbb{E}(d_-^2)$ simultaneously maximizing $\mathbb{E}(d_+^2)$. This can be achieved by minimizing the ratio $\mathbb{E}(d_-^2) / \mathbb{E}(d_+^2)$, which is solved by \emph{linear discriminant analysis} (LDA). However, we unfavor this approach as it does not allow control over the tradeoff between sensitivity and specificity. Instead, we propose to minimize the difference
\begin{eqnarray}
\lefteqn{(1-\alpha)\mathbb{E}(d_+^2) - \alpha \mathbb{E}(d_-^2) = } \nonumber\\
&& \tr( \bm{A} ((1-\alpha)\bm{C}_+  - \alpha\bm{C}_-) \bm{A}^\Tr ) = \tr( \bm{A} \bm{D}_\alpha \bm{A}^\Tr ), ~~~
\label{eq:difference}
\end{eqnarray}
where $0 \le \alpha \le 1$ controls the said tradeoff, and $\bm{D}_\alpha$ denotes the difference between the positive and the negative covariance matrices.

Note that since the scale of $\bm{A}$ is arbitrary, a trivial solution can be obtained. Even when fixing the scale, the solution will be a rank-$1$ matrix corresponding to the smallest eigenvector of $\bm{D}_\alpha$.
While this can be avoided by arbitrarily demanding orthonormality of $\bm{A}$, such a remedy is completely artificial.

Instead, we remind that one of the desired properties of a descriptor was \emph{efficiency}. In an efficient descriptor, each dimension should be statistically independent of the others. Replacing statistical independence by the more tractable lack of correlation, we demand
\begin{eqnarray}
\bm{I} = \mathbb{E}(\bm{p} \bm{p}^\Tr) = \bm{A} \mathbb{E}(\bm{g} \bm{g}^\Tr) \bm{A}^\Tr = \bm{A} \bm{C} \bm{A}^\Tr
\label{eq:efficiency}
\end{eqnarray}
where expectations are taken over all geometry vectors, and $\bm{C}$ denotes the covariance matrix of $\bm{g}$.

Combining (\ref{eq:difference}) with (\ref{eq:efficiency}), we obtain the following minimization problem
\begin{eqnarray}
\min_{\bm{A}} \tr( \bm{A} \bm{D}_\alpha \bm{A}^\Tr ) ~~\mathrm{s.t}~~ \bm{A} \bm{C} \bm{A}^\Tr = \bm{I},
\label{eq:optprob}
\end{eqnarray}
which we solve for an $n \times m$ matrix $\bm{A}$. The problem has a closed-form algebraic solution, which is easy to derive using variable substitution.
Since $\bm{C}$ is a positive-definite matrix, we can substitute $\bm{B} = \bm{A} \bm{C}^{1/2}$, obtaining an equivalent minimization problem
\begin{eqnarray}
\min_{\bm{B}} \tr( \bm{B} \bm{C}^{-1/2}\bm{D}_\alpha \bm{C}^{-1/2} \bm{B}^\Tr ) ~~\mathrm{s.t}~~ \bm{B} \bm{B}^\Tr = \bm{I},
\label{eq:optprob1}
\end{eqnarray}
($\bm{C}$ is symmetric and so is its root; we therefore keep writing $\bm{C}^{-1/2}$ instead of its transpose).
Let us denote by $\bm{C}^{-1/2}\bm{D}_\alpha \bm{C}^{-1/2} = \bm{U}\bm{\Lambda}\bm{U}^\Tr$ the eigendecomposition of the scaled covariance difference,
with the eigenvalues $\bm{\Lambda} = \mathrm{diag}(\lambda_1,\dots,\lambda_m)$ sorted in ascending order, and the corresponding orthonormal eigenvectors $\bm{U} = (\bm{u}_1,\dots,\bm{u}_m)$.
The solution to (\ref{eq:optprob1}) is given by the first $n$ smallest eigenvectors, $\bm{B} = \bm{U}_n^\Tr = (\bm{u}_1,\dots,\bm{u}_n)^\Tr$.
Note that one must ensure that all the eigenvectors correspond to negative eigenvalues; if this is not the case, $n$ has to be reduced.
%
Finally, the solution to our original problem (\ref{eq:optprob}) follows straightforwardly as
\begin{eqnarray}
\bm{A} = \bm{U}_n^\Tr \bm{C}^{-1/2}.
\end{eqnarray}

\subsection{Training set}
\label{sec:trainingset}

So far, we have described a learning scheme allowing to construct efficient spectral descriptors with uncorrelated elements based on covariances of geometry vectors describing positive and negative pairs of points. Having no practical possibility to model the statistics of these vectors, their covariance matrices have to be computed empirically from a training set of positive and negative examples. The construction of such a set is therefore crucial for obtaining a good descriptor.
In what follows, we describe how to construct the training set in order to achieve each of the desired properties mentioned before.

\smallskip

\noindent{\bf Localization.~} Let $x$ be a point on a training shape $X$. We fix a pair of radii $r < R$ and deem all points
$x_+ \in B_r(x)$ positive, while deeming negative all $x_- \in B^\mathrm{c}_R(x)$. Here, $B_r(x)$ denotes the geodesic metric ball of radius $r$ centered at $x$. Points lying in the ring $B_R(x) \setminus B_r(x)$ are excluded from both sets. If the shape possesses an intrinsic symmetry $\varphi : X \rightarrow X$, then $B_r(\varphi(x))$ is also included in the positive set, while $B_R(\varphi(x))$
is excluded from the negative set. The training set is created by sampling many reference points and corresponding positive and negative points on a collection of representative shapes.
The selection of $r$ and $R$ gives explicit control over the localization capability of the descriptor.

\smallskip

\noindent{\bf Discriminativity.~} Let $X$ and $X_-$ be knowingly dissimilar shapes (i.e., belonging to different classes we would like to tell apart). A random point $x$ on $X$ and a random point $x_-$ on $X_-$ are deemed negative. The training set is created by sampling many random pairs of points on knowingly dissimilar pairs of shapes.

\smallskip

\noindent{\bf Invariance.~} Let $X$ be a shape and $X_+$ its transformation belonging to a class of transformations invariance under which is desired. We further assume to be given a correspondence $\varphi : X \rightarrow X_+$ between the shapes. A random point $x$ on $X$ and the corresponding point $x_+ = \varphi(x)$ on $X_+$ are deemed positive. The training set is created by sampling many random points on a collection of null (reference) shapes, paired with corresponding points on the transformed versions of the null shape.

\smallskip

\noindent The combination of the positive and negative sets constructed this way allows to train for descriptor localization, discriminativity, and invariance properties.

\subsection{Sensitivity-Specificity tradeoff}
\label{sec:alpha}

The proposed learning scheme allows simple control over the tradeoff between the sensitivity and the specificity of the descriptor through the parameter $\alpha$.
The bigger is $\alpha$, the bigger is the relative influence of $\bm{C}_-$ compared to $\bm{C}_+$. Therefore, for large values of $\alpha$, the descriptor will emphasize producing large distances on the negative set (low false positive rate), while trying to keep small distances on the positive set (high true positive rate). As the result, high sensitivity is obtained. For small values of $\alpha$, the converse is observed: the descriptor emphasizes performance on the positive set, resulting in higher specificity.

\begin{figure}[tbp]
    \begin{center}
   \includegraphics[width=1\columnwidth]{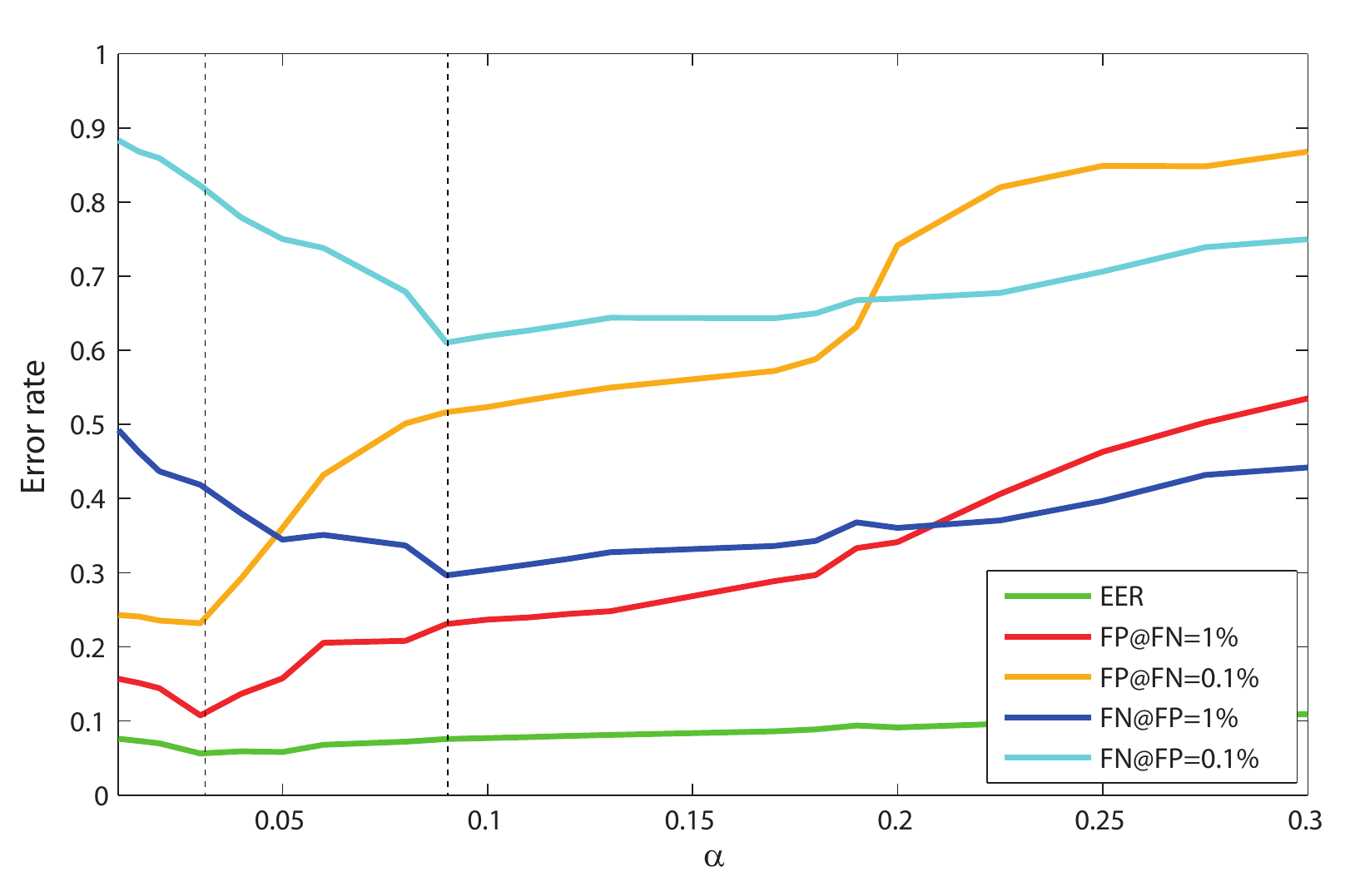}
   \caption{\label{fig:alpha} Error rates as a function of the parameter $\alpha$. Large values of $\alpha$ result in high sensitivity, while for small values high specificity is obtained.  } 
    \end{center}
\end{figure}

In order to select the optimal $\alpha$ for a highly-sensitive descriptor, we empirically compute the false negative rate at some small fixed false positive rate (e.g., $1\%$ or $0.1\%$) and select the $\alpha$ for which it is minimized. For highly-specific descriptors, $\alpha$ is selected to minimize the false positive rate at some small false negative rate. The behavior of the error rates as a function of $\alpha$ is illustrated in Figure~\ref{fig:alpha}.

\section{Experimental results}

The reported experiments were performed on the SHREC'10 robust correspondence benchmark \cite{bronstein2010shrec_corr}. The benchmark contains three distinct shape classes (human, dog, and horse), each shape undergoing ten different transformations (isometry, topology, sampling, global scaling, local scaling, holes, micro holes, Gaussian noise, and shot noise) with five strengths per transformation (from mild to very strong). Shapes are represented as triangular meshes with about $5 \times 10^4$ vertices (except for the sampling transformations, where the meshes are progressively decimated down to about $2.5 \times 10^3$ vertices). The benchmark also contains vertex-wise correspondences between the transformed shapes and the reference (null) shapes, including intrinsic bilateral symmetries. In all experiments, training was performed on the isometry, topology, and Gaussian noise transformations of the horse shape. As the negatives, we used five distinct meshes not included in the benchmark. For evaluation, we used the isometry, topology, holes, Gaussian noise and sampling transformations of the human shape, and the dog shape as the negative. All transformation strengths were used both for training and testing.

We used the finite elements scheme \cite{reuter2006laplace} to compute the first $300$ eigenvalues and eigenvectors of the Laplace-Beltrami operator on each shape. Neumann boundary conditions were used. The range of frequencies $\nu_\mathrm{max}$ was set to the $95$-percentile of $\nu_{300}$ over the entire set of training shapes. The interval was evenly divided into $m=150$ segments and the cubic spline basis was used as $\{b_j(\nu)\}$.
The training set containing $2.5\times 10^6$ $150$-dimensional triplets of the form $(\bm{g},\bm{g}_+,\bm{g}_-)$ was generated as described in Section~\ref{sec:trainingset} with $10^4$ negative examples per reference point. The radii $r$ and $R$ were set to $2\%$ and $5\%$ of the shape intrinsic diameter, respectively.
The parameter $\alpha$ was selected as described in Section~\ref{sec:alpha}. The values maximizing the descriptor specificity  and sensitivity were found to be $\alpha = 0.03$ and $0.09$, respectively (Figure~\ref{fig:alpha}). Two corresponding $12$-dimensional descriptors were trained. Examples of the obtained responses are shown in Figure~\ref{fig:kernels} (bottom).

\subsection{Descriptor performance}

Descriptor performance was tested on a distinct set of $2.5 \times 10^6$ triplets of points constructed in the same was as the training set but on different shapes.
For comparison, we also computed twelve-dimensional HKS and WKS descriptors. The HKS time scales were optimized according to \cite{bronstein2011shape}. The WKS energy levels and the variance $\sigma^2$ were set as described in \cite{aubry2011wave}. For the fairness of comparison, Euclidean distance was used for all descriptors.
Figure~\ref{fig:roc} shows the ROC curves of the compared descriptors in the low false positive and low false negatives work points. As argued before, the HKS is characterized by better performance over the WKS at low false negative rates, while the WKS outperforms the HKS in the low positive rates range. The trained descriptors significantly outperform both the HKS and the WKS in the low false negative rates range, with almost a $40\%$ increase in the true negative rate at $FN=0.1\%$. The trained high-sensitivity descriptor outperforms WKS by about $6\%$ true positive rate at $FP=1\%$. The improvement becomes more modest at $FN=0.1\%$.

\subsection{Localization}

In order to visualize the localization capability of different descriptors, a reference point was selected on the human shape. The distance between the descriptor at that point was computed to the rest of the points on that shape, to the points of an approximate isometry of the human shape, and to the points on the dog shape.
Figure~\ref{fig:distance} visualizes these normalized distances on a common scale. We observe poor localization capabilities of the HKS along with exceptional localization power of the WKS. The trained high-sensitivity descriptor exhibits even better localization.
Both the HKS and the WKS confuse between the reference point on the man's foot and a region on his hand fingers, which have similar geometric content. On the other hand, our descriptor does not make this confusion. We remind that in the training set, for every reference point all points except its small neighborhood were included as negatives. Even though a different shape was used during the training, the descriptor still seems to be capable of generalizing these relationships.

\begin{figure}[tbp]
    \begin{center}
   \includegraphics[width=1\columnwidth]{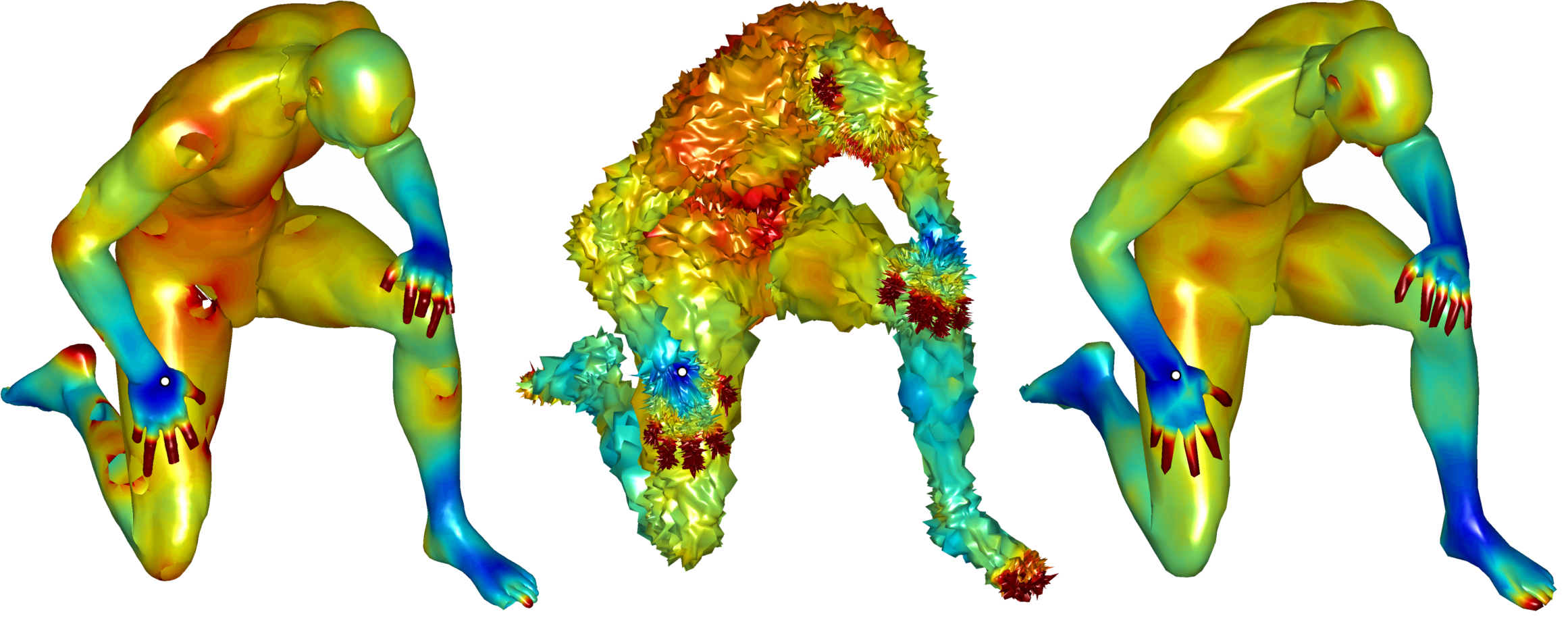}
   \caption{\label{fig:xforms} Normalized Euclidean distance between the descriptor at a reference point on the right hand (white dot) and descriptors computed at rest of the points of the same shape for a twelve-dimensional trained optimal descriptors. Left-to-right: holes, Gaussian noise, and sampling transformations from the SHREC'10 benchmark. } 
    \end{center}
\end{figure}

Finally, both the HKS and the WKS find many points on the dog shape that resemble the reference point on the man's foot. Our descriptor does not make this confusion as it was trained for discriminativity with numerous negative examples from distinct shapes.
Figure~\ref{fig:xforms} shows additional examples of distances computed on other transformations of the human shape using the trained descriptor. In all cases, good localization is observed.

\subsection{Correspondence}

\begin{figure}[tbp]
    \begin{center}
   \includegraphics[width=1\columnwidth]{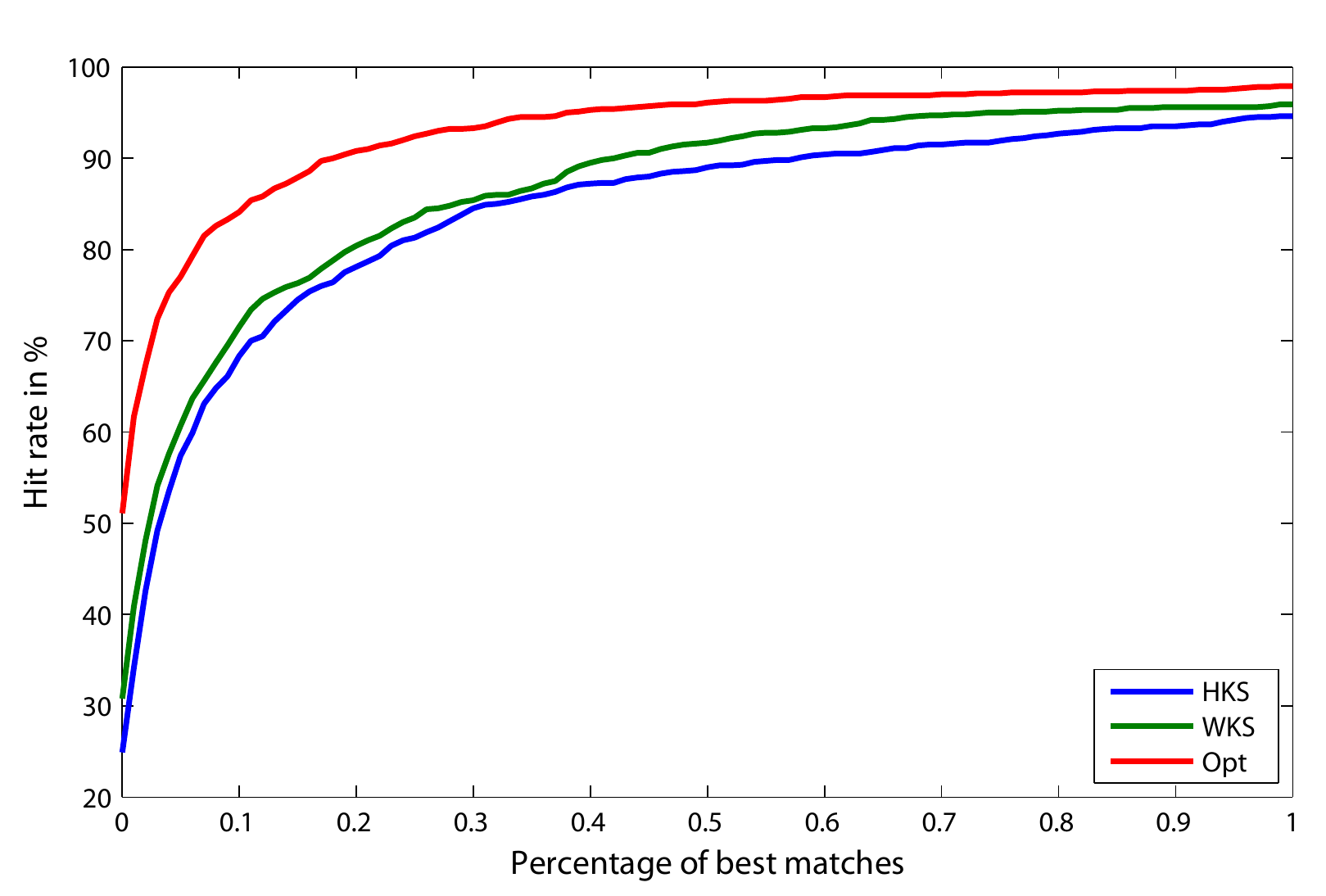}
   \caption{\label{fig:hitrate} CMC curve showing the percentage of correct correspondences found in a subset of the first best matches (up to $1\%$ of total points) using different spectral descriptors.  }
    \end{center}
\end{figure}

While evaluation of a particular descriptor-based correspondence algorithms is beyond the scope of this paper,
in order to test the performance of the trained high-selectivity descriptor in shape matching tasks, we performed an experiment similar to \cite{aubry2011wave}. $1000$ reference points were sampled on the human shape using farthest point sampling in the descriptor space. Such points coincided well with visually ``interesting'' features. Each reference point was matched to all the points on the transformed versions of the shape. We computed the probability of finding the correct match (including the symmetric one) within the first $k$ best matches. The CMC curve in Figure~\ref{fig:hitrate} depicts the hit rate of different descriptors for up to about first $500$ matches (corresponding to $1\%$ of the total points on the shape). The trained descriptor significantly outperforms both the HKS and the WKS. In fact, our descriptor returns the first correct match with over $50\%$ probability, compared to about $25\%$ and $30\%$ in the case of HKS and WKS, respectively. 

While the WKS consistently outperforms the HKS on this matching task, we did not notice the dramatic difference reported in \cite{aubry2011wave}. A possible explanation can be the fact that we used only $12$ dimensions, while the authors of \cite{aubry2011wave} used a higher-dimensional descriptor. Another, more probable, reason is the fact that in all our experiments Euclidean distance was used as the dissimilarity between the descriptors, while in \cite{aubry2011wave} the authors used WKS with the normalized $L^1$ distance.
We defer to future studies the treatment of distances other than $L^2$; however, we believe that for the fairness of comparison the same distance must be used for all descriptors.

\section{Conclusion}

We presented a generic framework for the construction of feature descriptors for deformable shapes
based on their spectral properties. The proposed descriptor is computed by applying a bank of ``filters''
to the shape's geometric features at different ``frequencies'', and it generalizes the heat and wave kernel signatures. 
We also showed a learning approach allowing to construct optimal filters for specific shape analysis tasks, resembling in its spirit optimal signal filtering by means of a Wiener filter.

We formulated the learning approach in terms of the $L^2$ distance and related it to Mahalanobis metric learning. While the adopted algebraic solution gave good results, other Mahalanobis metric learning approaches, such as the maximum-margin learning \cite{weinberger2006distance} can be readily used. Some of these metric learning approaches were designed with a specific task in mind (e.g., ranking), and might be beneficial for the construction of spectral descriptors in some applications. Evidence shows that distances other than the Euclidean one (e.g., the $L^1$ distance) improve the performance of spectral descriptors. Also, applications where compact and easily searchable descriptors are of importance may benefit from hash learning techniques \cite{weiss2008spectral}, essentially based on the Hamming distance. We intend to explore alternative learning frameworks and different distances in follow-up studies.

While the main focus of this paper was the construction of the descriptor itself, in future studies we are going to explore its performance in real shape retrieval and matching tasks. 
Particularly, in retrieval tasks spectral feature descriptors are used to generate global shape descriptors by means of vector quantization or sparse coding, a growingly popular alternative in the computer vision community. Taking this highly non-linear process into account when constructing the feature descriptor will also be a subject of our future research.

\bibliographystyle{IEEEtran}
\bibliography{descriptors}

\begin{thebibliography}{10}
\providecommand{\url}[1]{#1}
\csname url@samestyle\endcsname
\providecommand{\newblock}{\relax}
\providecommand{\bibinfo}[2]{#2}
\providecommand{\BIBentrySTDinterwordspacing}{\spaceskip=0pt\relax}
\providecommand{\BIBentryALTinterwordstretchfactor}{4}
\providecommand{\BIBentryALTinterwordspacing}{\spaceskip=\fontdimen2\font plus
\BIBentryALTinterwordstretchfactor\fontdimen3\font minus
  \fontdimen4\font\relax}
\providecommand{\BIBforeignlanguage}[2]{{%
\expandafter\ifx\csname l@#1\endcsname\relax
\typeout{** WARNING: IEEEtran.bst: No hyphenation pattern has been}%
\typeout{** loaded for the language `#1'. Using the pattern for}%
\typeout{** the default language instead.}%
\else
\language=\csname l@#1\endcsname
\fi
#2}}
\providecommand{\BIBdecl}{\relax}
\BIBdecl

\bibitem{gelfand2005rgr}
N.~Gelfand, N.~J. Mitra, L.~J. Guibas, and H.~Pottmann, ``Robust global
  registration,'' in \emph{Proc. SGP}, 2005.

\bibitem{wang2011matching}
C.~Wang, A.~M. Bronstein, M.~M. Bronstein, and N.~Paragios, ``{Discrete minimum
  distortion correspondence problems for non-rigid shape matching},'' in
  \emph{Proc. Scale Space and Variational Methods (SSVM)}, 2011.

\bibitem{mitra2006pfs}
N.~J. Mitra, L.~J. Guibas, J.~Giesen, and M.~Pauly, ``Probabilistic
  fingerprints for shapes,'' in \emph{Proc. SGP}, 2006.

\bibitem{bronstein2011shape}
A.~Bronstein, M.~Bronstein, L.~Guibas, and M.~Ovsjanikov, ``Shape google:
  geometric words and expressions for invariant shape retrieval,'' \emph{ACM
  Transactions on Graphics (TOG)}, vol.~30, no.~1, p.~1, 2011.

\bibitem{skraba2010persistence}
P.~Skraba, M.~Ovsjanikov, F.~Chazal, and L.~Guibas, ``{Persistence-based
  segmentation of deformable shapes},'' in \emph{Proc. NORDIA}, 2010, pp.
  45--52.

\bibitem{belongie2000shape}
S.~Belongie, J.~Malik, and J.~Puzicha, ``Shape context: A new descriptor for
  shape matching and object recognition,'' in \emph{Proc. NIPS}, 2000.

\bibitem{johnson1999usi}
A.~E. Johnson and M.~Hebert, ``{Using spin images for efficient object
  recognition in cluttered 3D scenes},'' \emph{Trans. PAMI}, vol.~21, no.~5,
  pp. 433--449, 1999.

\bibitem{manay2004iis}
S.~Manay, B.~Hong, A.~Yezzi, and S.~Soatto, ``{Integral invariant
  signatures},'' \emph{Lecture Notes in Computer Science}, pp. 87--99, 2004.

\bibitem{pauly2003msf}
M.~Pauly, R.~Keiser, and M.~Gross, ``{Multi-scale feature extraction on
  point-sampled surfaces},'' in \emph{Computer Graphics Forum}, vol.~22, no.~3,
  2003, pp. 281--289.

\bibitem{hamza2003geodesic}
A.~Hamza and H.~Krim, ``Geodesic object representation and recognition,'' in
  \emph{Discrete Geometry for Computer Imagery}, 2003, pp. 378--387.

\bibitem{elad2003bending}
A.~Elad and R.~Kimmel, ``On bending invariant signatures for surfaces,''
  \emph{IEEE Trans. Pattern Analysis and Machine Intelligence}, pp. 1285--1311,
  2003.

\bibitem{lipman2009mobius}
Y.~Lipman and T.~Funkhouser, ``M{\"o}bius voting for surface correspondence,''
  in \emph{ACM Trans. on Graphics}, vol.~28, no.~3, 2009, p.~72.

\bibitem{berard1994embedding}
P.~B{\'e}rard, G.~Besson, and S.~Gallot, ``{Embedding Riemannian manifolds by
  their heat kernel},'' \emph{Geometric and Functional Analysis}, vol.~4,
  no.~4, pp. 373--398, 1994.

\bibitem{coifman2006diffusion}
R.~Coifman and S.~Lafon, ``Diffusion maps,'' \emph{Applied and Computational
  Harmonic Analysis}, vol.~21, no.~1, pp. 5--30, 2006.

\bibitem{memoli2009spectral}
F.~M{\'e}moli, ``Spectral {G}romov-{W}asserstein distances for shape
  matching,'' in \emph{Proc. ICCV Workshops}, 2009, pp. 256--263.

\bibitem{bronstein2010gromov}
A.~Bronstein, M.~Bronstein, R.~Kimmel, M.~Mahmoudi, and G.~Sapiro, ``A
  {G}romov-{H}ausdorff framework with diffusion geometry for
  topologically-robust non-rigid shape matching,'' \emph{Int'l Journal of
  Computer Vision}, vol.~89, no.~2, pp. 266--286, 2010.

\bibitem{lévy2006laplace}
B.~L{\'e}vy, ``{Laplace-Beltrami eigenfunctions towards an algorithm that
  understands geometry},'' in \emph{Proc. SMI}, 2006, pp. 13--13.

\bibitem{rustamov2007laplace}
R.~Rustamov, ``{Laplace-Beltrami eigenfunctions for deformation invariant shape
  representation},'' in \emph{Proc. Symp. on Geometry Processing (SGP)}, 2007,
  pp. 225--233.

\bibitem{sun2009concise}
J.~Sun, M.~Ovsjanikov, and L.~Guibas, ``{A Concise and Provably Informative
  Multi-Scale Signature Based on Heat Diffusion},'' in \emph{Computer Graphics
  Forum}, vol.~28, no.~5, 2009, pp. 1383--1392.

\bibitem{aubry2011wave}
M.~Aubry, U.~Schlickewei, and D.~Cremers, ``The wave kernel signature-a quantum
  mechanical approach to shape analyis,'' in \emph{Proc. CVPR}, 2011.

\bibitem{bronstein2010shrec_corr}
A.~Bronstein, M.~Bronstein, U.~Castellani, A.~Dubrovina, L.~Guibas, R.~Horaud,
  R.~Kimmel, D.~Knossow, E.~von Lavante, D.~Mateus \emph{et~al.}, ``{SHREC
  2010: robust correspondence benchmark},'' 2010.

\bibitem{bronstein2010shrec_ret}
A.~Bronstein, M.~Bronstein, U.~Castellani, B.~Falcidieno, A.~Fusiello,
  A.~Godil, L.~Guibas, I.~Kokkinos, Z.~Lian, M.~Ovsjanikov \emph{et~al.},
  ``{SHREC 2010: robust large-scale shape retrieval benchmark},'' 2010.

\bibitem{aflalo2011wiener}
J.~Aflalo, A.~M. Bronstein, M.~M. Bronstein, and R.~Kimmel, ``{Deformable shape
  retrieval by learning diffusion kernels},'' in \emph{Proc. Scale Space and
  Variational Methods (SSVM)}, 2011.

\bibitem{kac1966can}
M.~Kac, ``Can one hear the shape of a drum?'' \emph{The American Mathematical
  Monthly}, vol.~73, no.~4, pp. 1--23, 1966.

\bibitem{reuter2006laplace}
M.~Reuter, F.~Wolter, and N.~Peinecke, ``{Laplace-Beltrami spectra as
  ``Shape-DNA'' of surfaces and solids},'' \emph{Computer-Aided Design},
  vol.~38, no.~4, pp. 342--366, 2006.

\bibitem{sharma2010shape}
A.~Sharma and R.~Horaud, ``Shape matching based on diffusion embedding and on
  mutual isometric consistency,'' in \emph{Proc. CVPR Workshops}, 2010, pp.
  29--36.

\bibitem{Iasonas}
M.~M. Bronstein and I.~Kokkinos, ``Scale-invariant heat kernel signatures for
  non-rigid shape recognition,'' in \emph{Proc. CVPR}, 2010.

\bibitem{raviv2010volumetric}
D.~Raviv, M.~Bronstein, A.~Bronstein, and R.~Kimmel, ``Volumetric heat kernel
  signatures,'' in \emph{Proc. ACM Workshop on 3D Object Retrieval}, 2010, pp.
  39--44.

\bibitem{wiener1964extrapolation}
N.~Wiener, \emph{Extrapolation, interpolation, and smoothing of stationary time
  series}.\hskip 1em plus 0.5em minus 0.4em\relax Wiley.

\bibitem{yang2006distance}
L.~Yang and R.~Jin, ``Distance metric learning: A comprehensive survey,''
  \emph{Michigan State Universiy}, pp. 1--51, 2006.

\bibitem{weinberger2006distance}
K.~Weinberger, J.~Blitzer, and L.~Saul, ``Distance metric learning for large
  margin nearest neighbor classification,'' in \emph{Proc. NIPS}, 2006.

\bibitem{davis2007information}
J.~Davis, B.~Kulis, P.~Jain, S.~Sra, and I.~Dhillon, ``Information-theoretic
  metric learning,'' in \emph{Proc. ICML}, 2007, pp. 209--216.

\bibitem{strecha2011ldahash}
C.~Strecha, A.~Bronstein, M.~Bronstein, and P.~Fua, ``{LDAHash}: Improved
  matching with smaller descriptors,'' \emph{IEEE Trans. Pattern Analysis and
  Machine Intelligence}, 2011.

\bibitem{weiss2008spectral}
Y.~Weiss, A.~Torralba, and R.~Fergus, ``Spectral hashing,'' in \emph{Proc.
  NIPS}, 2008.

\end{thebibliography}

\end{document}